\documentclass[letterpaper]{article} 
\usepackage{aaai24}  
\usepackage{times}  
\usepackage{helvet}  
\usepackage{courier}  
\usepackage[hyphens]{url}  
\usepackage{graphicx} 
\usepackage{natbib}  
\usepackage{caption} 
\usepackage{algorithm}
\usepackage{algorithmic}
\UseRawInputEncoding
\usepackage{multirow}
\usepackage{booktabs}
\usepackage{pifont}
\usepackage{bbding}
\usepackage{subcaption}
\usepackage{newfloat}
\usepackage{listings}
\DeclareCaptionStyle{ruled}{labelfont=normalfont,labelsep=colon,strut=off} 
\floatstyle{ruled}
\newfloat{listing}{tb}{lst}{}
\floatname{listing}{Listing}
\title{SciEval: A Multi-Level Large Language Model \\ Evaluation Benchmark for Scientific Research}
\author{
    Liangtai Sun, Yang Han, Zihan Zhao, Da Ma, Zhennan Shen, Baocai Chen, Lu Chen\footnote{The corresponding authors are Lu Chen and Kai Yu.}, Kai Yu\footnotemark[1]
}

\affiliations {
    X-LANCE Lab, Department of Computer Science and Engineering\\
    MoE Key Lab of Artificial Intelligence, AI Institute, Shanghai Jiao Tong University\\
    Shanghai Jiao Tong University, Shanghai, China \\
    \{slt19990817, csyanghan, zhao\_mengxin, mada123\}@sjtu.edu.cn \\
    \{ieee-szn, 15368493547, chenlusz, kai.yu\}@sjtu.edu.cn

}

\usepackage{bibentry}

\begin{document}

\maketitle

\begin{abstract}

Recently, there has been growing interest in using Large Language Models (LLMs) for scientific research. Numerous benchmarks have been proposed to evaluate the ability of LLMs for scientific research. However, current benchmarks are mostly based on pre-collected objective questions. This design suffers from data leakage problem and lacks the evaluation of subjective Q/A ability. In this paper, we propose {\em SciEval}, a comprehensive and multi-disciplinary evaluation benchmark to address these issues. Based on Bloom's taxonomy, {\em SciEval} covers four dimensions to systematically evaluate scientific research ability. In particular, we design a ``dynamic" subset based on scientific principles to prevent evaluation from potential data leakage. Both objective and subjective questions are included in {\em SciEval}. These characteristics make {\em SciEval} a more effective benchmark for scientific research ability evaluation of LLMs. Comprehensive experiments on most advanced LLMs show that, although GPT-4 achieves SOTA performance compared to other LLMs, there is still substantial room for improvement, especially for dynamic questions. The codes and data are publicly available on \url{https://github.com/OpenDFM/SciEval}.

\end{abstract}

\section{Introduction}

Large Language Models (LLMs), such as ChatGPT~\cite{schulman2022chatgpt}, have attracted widespread attention in general scenarios, including information search, code generation, and more. In the field of science, LLMs have also shown preliminary potential in improving scientific research efficiency and transforming scientific research paradigms~\cite{blanco2023role,wang2023novel}. In the meanwhile, several scientific LLMs have been proposed by researchers~\cite{GALACTICA,luo2022biogpt,frey2022neural}. In the general field, there are already numerous evaluation benchmarks to evaluate the language understanding, language generation and reasoning capabilities of LLMs, such as MMLU~\cite{hendrycks2020measuring}, AGIEval~\cite{zhong2023agieval}, and C-EVAL~\cite{huang2023c}, shown in Table \ref{tab:dataset_compare}. Although these benchmarks cover data of science domain, the data sources are usually confined to educational materials, which can not adequately assess the research ability of LLMs and not align with real-life scientific research scenarios. In addition, some benchmarks have been proposed to evaluate the scientific capability of LLMs, such as MultiMedQA~\cite{singhal2023large}, ChemLLMBench~\cite{guo2023indeed}, and MATH~\cite{hendrycks2021measuringmath}, while these benchmarks are restricted to a specific scientific discipline, leaving a lack of a more general scientific evaluation benchmark.\footnote{Due to the page limitation, we only compare some widely used benchmarks. For more information, we refer to~\cite{chang2023survey}.} In addition, these benchmarks (1) lack evaluation systems for scientific capabilities, (2) are all based on objective questions, which are insufficient to assess scientific abilities, and (3) face the risk of data leakage.

\begin{table*}[h]
\centering
\begin{tabular}{lllllcl}
\toprule
\textbf{Name} & \textbf{Category}                                                                  & \textbf{Ability} & \textbf{Source}                                                                                  & \textbf{Data Type}  & \textbf{Dynamic} & \textbf{\#Data}  \\ \midrule
 MMLU          & \begin{tabular}[c]{@{}l@{}}humanities, social \\ science, STEM, other\end{tabular} & BK, KA, SC       & exam, book, course                                                                               & objective  & \XSolidBrush  & 14079       \\\cmidrule{2-7}   AGIEval       & social science, STEM                                                               & BK, KA, SC       & exam                                                                                             & objective & \XSolidBrush  & 8062         \\ \cmidrule{2-7}  C-EVAL        & \begin{tabular}[c]{@{}l@{}}humanities, social \\ science, STEM, other\end{tabular} & BK, KA, SC       & exam                                                                                              & objective  & \XSolidBrush  & 12342      \\ \midrule
MultiMedQA    & medical                                                                            & BK, KA, RA       & exam, research     & objective & \XSolidBrush & 13115      \\ \cmidrule{2-7} ChemLLMBench  & chemistry                                                                          & BK,KA            & knowledge base                                                                                     & objective   & \XSolidBrush   & 800     \\ \cmidrule{2-7} MATH          & mathematics                                                                       & SC               & exam                                                                                               & objective  & \XSolidBrush & 5000         \\ \cmidrule{1-7} SciEval   & science                                                                            & BK, KA,SC, RA    & \begin{tabular}[c]{@{}l@{}}community QA, \\ knowledge base\end{tabular}                            & \begin{tabular}[c]{@{}l@{}}objective + \\ subjective \end{tabular} & \CheckmarkBold  & 15901      \\ \bottomrule
\end{tabular}
\caption{Dataset comparison of SciEval and some other datasets covering science domain.``BK" stands for Basic Knowledge, ``KA" stands for Knowledge Application, ``SC" stands for Scientific Calculation, and ``RA" stands for Research Ability.}
\label{tab:dataset_compare}
\end{table*}

In response to this gap, we present SciEval, an English benchmark designed to evaluate advanced abilities of LLMs in the scientific domain. SciEval consists of a total of about 18000 challenging scientific questions, spanning three important basic science fields: chemistry, physics and biology, each of which is further divided into multiple sub-topics. SciEval mainly has the following three characteristics:
\begin{itemize}
    \item \textbf{Multi-level and comprehensive evaluation of the ability of LLMs in the scientific field.} Scientific ability of LLMs needs to be evaluated from multiple aspects. Leveraging cognitive domains of Bloom's taxonomy~\cite{krathwohl2002revision,forehand2010bloom}, which covers six levels, SciEval evaluates the scientific capabilities of large language models across four dimensions: basic knowledge, knowledge application, scientific calculation, and research ability, where each capability aligns with one or more cognitive levels.
    \item \textbf{Combination of objective and subjective questions.} SciEval is mainly based on objective questions, which allow for quick and standard model evaluations, involving multiple-choice, fill-in-the-blank, and judgment questions. These questions can help us understand whether the model can correctly understand and memorize scientific knowledge. However, objective questions are insufficient to assess scientific capability holistically.
    To better assess scientific reasoning and application ability, SciEval introduces a small number of subjective questions, involving a total of twelve basic science experiments, which is named Experimental Data. 
    \item \textbf{Dynamic data generation based on basic scientific principles.} The huge amount of training data used for pre-training LLMs may cause the risk of data leakage for evaluation. In order to solve this problem, one of the main features of SciEval is the use of Dynamic Data, which can prevent potential data leakage and ensure the fairness and credibility of the evaluation results. The Dynamic Data will be updated regularly, and we will maintain a stable version to make a fair comparison of model performance. And the objective questions other than Dynamic Data are referred to as Static Data.
\end{itemize}


We conduct experiments to evaluate LLMs on SciEval in answer-only, chain-of-thought and few-shot settings. Results indicate that GPT-4 is the strongest model, with only GPT-4, GPT-3.5-turbo and Claude-v1.3 surpassing 60\% average accuracy on Static Data, signifying considerable opportunities for improvement. With the results of Dynamic Data, we find that these LLMs have little knowledge about molecules, and most models could only retain near-random accuracy in the physics subset. As for Experimental Data, some top-tier models could perform satisfactorily in experimental principle and design, while almost all models struggle to analyze the experimental results. With the analysis of experiment results, we claim that training on large-scale scientific corpus is helpful for the scientific ability of LLMs, and most LLMs perform bad on calculation problems, especially in physics domain. We hope SciEval can provide an excellent benchmark for the assessment of scientific capability of LLMs, and promote wide application in science.


\section{Related Work}



\subsection{General Benchmarks for LLMs}

To evaluate the performance of LLMs across different tasks, several benchmarks have been proposed.
MMLU~\cite{hendrycks2020measuring} aims to develop a comprehensive test for evaluating text models in multi-task contexts. 
HELM~\cite{liang2022holistic} offers a comprehensive assessment, evaluating LLMs across various aspects, such as language understanding and common-sense reasoning.
Big-Bench~\cite{srivastava2022beyond} introduces 204 challenging tasks covering various domains, aiming to evaluate tasks beyond the capabilities of existing language models.
AGIEval~\cite{zhong2023agieval} serves as an evaluation framework for assessing the performance of foundation models in human-centric standardized exams. 
C-Eval~\cite{huang2023c} assesses the advanced knowledge and reasoning capabilities of foundation models in Chinese.

\begin{figure*}[h]
    \centering
    \includegraphics[width=0.9\textwidth]{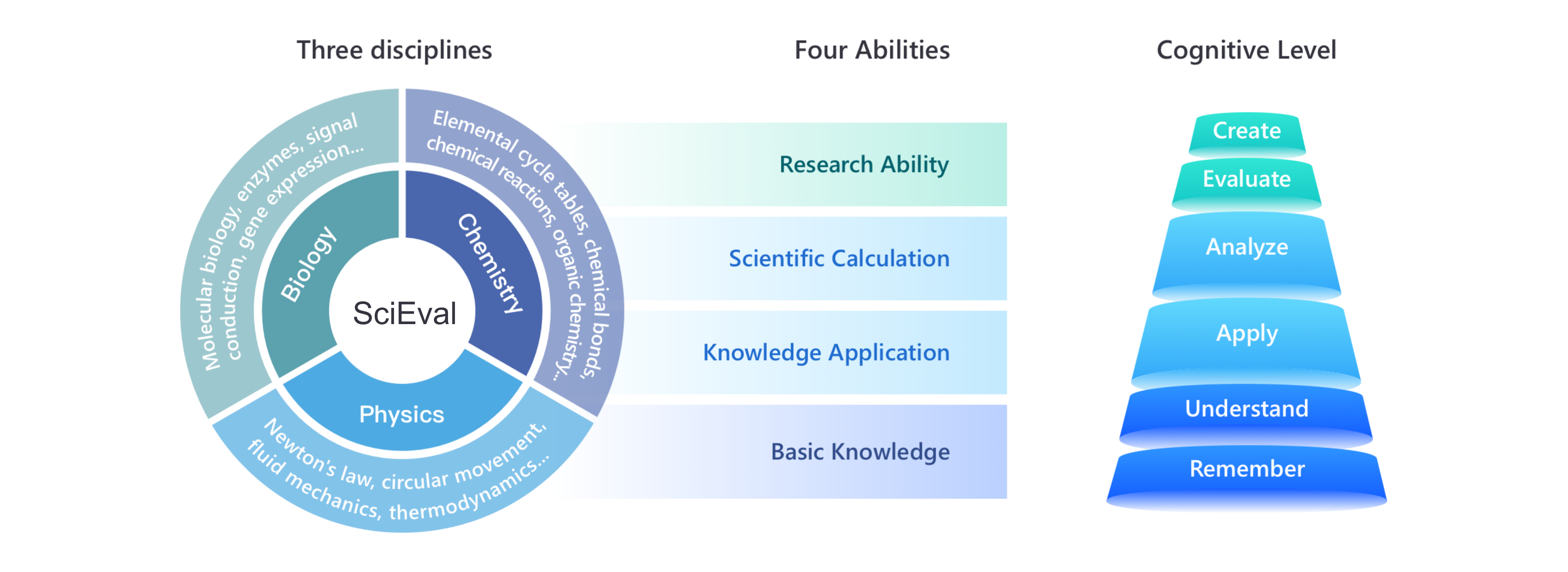}
    \caption{The illustration of the evaluation system. SciEval covers three disciplines with amounts of sub-topics, and investigates four abilities, corresponding to six cognitive levels.}
    \label{fig:system}
\end{figure*}

\subsection{Specific Benchmarks for LLMs}

Apart from general tasks, specific benchmarks are designed for certain downstream tasks. MultiMedQA~\cite{singhal2023large} focuses on medical question-answering, evaluating LLMs in terms of clinical knowledge and QA abilities.
MATH~\cite{hendrycks2021measuringmath} assesses reasoning and problem-solving proficiencies of LLMs in mathematics. 
ScienceQA~\cite{lu2022learn} proposes a multi-modal benchmark with a diverse set of science topics and annotations of their answers with corresponding lectures and explanations, collected from elementary and high school science curricula.
SCIBENCH~\cite{wang2023scibench} examines the reasoning capabilities required for complex scientific problem-solving and proposes two datasets of college-level scientific problems.
Compared to these benchmarks, SciEval (1) evaluates scientific capabilities from multiple aspects, having a broader coverage, (2) uses data of community Q\&A, which is more flexible and diverse, (3) designs a subset of dynamic data, making an effort to mitigate data leakage.

\section{The SciEval Dataset}


\subsection{Scientific Research Evaluation System}
\label{evaluation_system}

Scientific research requires different dimensions of knowledge, such as understanding and calculation, thence evaluation of scientific ability should be conducted at multiple levels. Bloom's taxonomy is a set of three hierarchical methods used for classification of educational learning objectives covering cognitive, affective and psychomotor domains. The cognitive domain is frequently used to structure curriculum learning objectives, assessments and activities, and is broken into six levels: Remember, Understand, Apply, Analyze, Evaluate and Create, as is shown in Figure \ref{fig:system}, which are suitable for the evaluation of scientific capability.

Based on the cognitive domain of Bloom's taxonomy, the evaluation system of SciEval consists of four knowledge dimensions: \textit{Basic Knowledge (BK), Knowledge Application (KA), Scientific Calculation (SC), and Research Ability (RA)}. As is shown in Figure \ref{fig:system}, BK primarily assesses the fundamental scientific knowledge of LLMs. KA focuses on how to apply basic knowledge to solve scientific problems, requiring models to have comprehension, application, and analysis abilities. SC is a specialized application of knowledge that further examines complex reasoning capabilities of LLMs based on their general knowledge application abilities. RA assesses evaluation capabilities at a higher cognitive level, requiring models to participate in various aspects of scientific research, including problem formulation, experimental design, data analysis, and summarization.

Based on the evaluation system, we design three different types of data: \textit{Static Data, Dynamic Data, and Experimental Data}. The Static Data covers all these four knowledge dimensions and will remain constant throughout, while the Dynamic Data examines from the aspects of Knowledge Application and Scientific Calculation and will be regularly updated to prevent any data leakage. The Experimental Data comprises a set of questions for twelve scientific experiments and can be used to evaluate the Research Ability.

\begin{figure*}[h]
    \centering
    \includegraphics[width=0.90\textwidth]{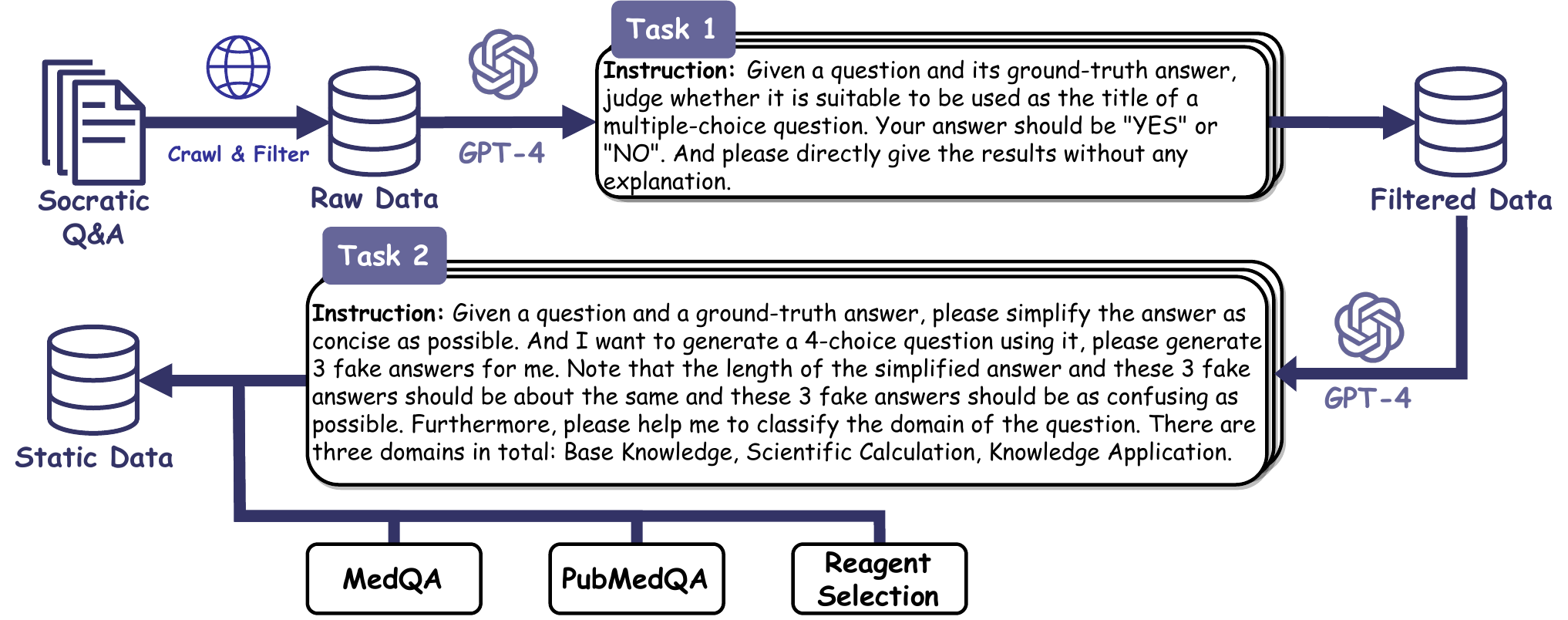}
    \caption{Data Collection steps of Static Data}
    \label{fig:dc_static}
\end{figure*}

\subsection{Data Collection}

\label{data_collection}

\subsubsection{Static Data}

The collection steps of Static Data are shown in Figure \ref{fig:dc_static}. The primary source of Static Data is Socratic Q\&A\footnote{\url{https://socratic.org}}, a community-driven website that covers a wide range of subjects such as science and literature. Specifically, we collect data from the fields of biology, chemistry, and physics. To ensure quality, we employ rule-based methods to preprocess the crawled data. While gathering the questions, we found that not all of them are suitable as titles. To address this, we utilize GPT-4 with the ``Task 1" prompt, as depicted in Figure \ref{fig:dc_static}, to process these questions. Since most of the collected questions are open-ended and challenging to evaluate, we employ GPT-4 to simplify ground-truth answers and generate three wrong answers to formulate them as multiple-choice questions. Additionally, we classify the questions into their respective knowledge domains. And during this process, we manually check the generated content of GPT-4 to ensure data quality.

To make the dataset more diverse and comprehensive, we further integrate data from some publicly available datasets:
\begin{itemize}
    \item \textbf{MedQA}~\cite{jin2021disease} is a free-form multiple-choice OpenQA dataset for solving medical problems, collected from professional medical board exams. We use the test set of USMLE, which is the English subset of MedQA.
    \item \textbf{PubMedQA}~\cite{jin2019pubmedqa} is a biomedical question-answering dataset collected from PubMed abstracts. The task of PubMedQA is to answer research questions with yes/no/maybe using the corresponding abstracts, which is fit for evaluating the literature comprehension ability. We incorporate 1000 expert-annotated data from it and frame them as judgment questions.
    \item \textbf{Reagent Selection}~\cite{guo2023indeed} involves the identification and proposal of the most fitting reagents for a specific chemical reaction or process, which is a subset of ChemLLMBench. We randomly select 40\% data and formulate them as multiple-choice questions.
\end{itemize}

\subsubsection{Dynamic Data}

The current training of LLMs often uses a large amount of data, resulting in a risk of data leakage for evaluation. In order to solve this problem, we design a ``dynamic" subset, which can generate data dynamically according to scientific principles. The dynamic subset covers two disciplines, chemistry and physics. For chemistry data, we use the basic information and properties of molecules crawled from PubChem\footnote{\url{https://pubchem.ncbi.nlm.nih.gov/}} to create data. For physics data, we manually write some Python scripts according to the physics formulas. When obtaining the evaluation dataset, we will provide a regenerated version to users and we will update it regularly, while at the same time, we will maintain a stable version of the dynamic data to make a fair comparison.

\subsubsection{Experimental Data}

To better evaluate the scientific thoughts and abilities of LLMs, SciEval introduces a subset of experimental data, involving 12 different basic scientific experiments. These experiments are collected from basic science experiment courses at university, and each experiment conducts a comprehensive investigation of the ability of LLMs in scientific research and experimentation from the perspectives of experimental principle, process, and analysis and summarization of experimental results.

\subsection{Data Statistics}

\label{data_statistics}

Summarized statistics are shown in Table \ref{tab:stat_sd}, where we only count Static Data. For Dynamic Data, the chemistry part examines the KA ability and contains 2000 data, while the physics part evaluates the SC ability and involves 890 data. All these questions are in English and we show some data examples in Appendix D.

\begin{table}[H]
\centering
\begin{tabular}{l|ccc}
\hline
Ability                & Bio & Chem & Phy \\ \hline
Basic Knowledge        & 2147    & 2914      & 456     \\
Knowledge Application  & 1379    & 3720      & 36      \\
Scientific Calculation & 301     & 3401      & 1165    \\
Research Ability       & 1000    & 0         & 0       \\
Total                  & 4830    & 10035     & 1657    \\ \hline
\end{tabular}
\caption{Statistics of Static Data}
\label{tab:stat_sd}
\end{table}

For Static Data, we further split the data into dev, valid, and test set. For each data source, each knowledge domain, and each discipline, we randomly select 5 data to form the dev set, which can be used for few-shot learning, and we split the remaining data with a ratio of 1:9 to construct the valid set and test set respectively. 


\section{Experiment}

\begin{table*}[h]
\centering
\begin{tabular}{lcccccc}
\toprule
\textbf{Model}      & \textbf{Creator} & \textbf{\#Parameters} & \textbf{Access} & \textbf{SD} & \textbf{DD} & \textbf{ED} \\ \midrule
GPT-4               & OpenAI           & \textit{undisclosed}  & API             &  \CheckmarkBold                    & \CheckmarkBold                      & \CheckmarkBold                           \\
GPT-3.5-turbo       & OpenAI           & \textit{undisclosed}  & API             & \CheckmarkBold                     & \CheckmarkBold                      & \CheckmarkBold                           \\
Claude-v1.3         & Anthropic        & \textit{undisclosed}  & API             & \CheckmarkBold                     &  \CheckmarkBold                     &  \CheckmarkBold                          \\
Claude-instant-v1.1 & Anthropic        & \textit{undisclosed}  & API             &  \CheckmarkBold                    &  \CheckmarkBold                     &  \CheckmarkBold                          \\
ERNIE Bot           & Baidu            & \textit{undisclosed}  & Web             &                      &                       &  \CheckmarkBold                          \\
SparkDesk           & iFLYTEK          & \textit{undisclosed}  & Web             &                      &                       &  \CheckmarkBold                          \\
Vicuna              & LMSYS            & 13B                   & Weights         & \CheckmarkBold                     &  \CheckmarkBold                     &                            \\
Galactica           & Meta             & 30B, 6.7B             & Weights         & \CheckmarkBold                     &   \CheckmarkBold                    &                            \\
ChatGLM2            & Tsinghua         & 6B                    & Weights         &  \CheckmarkBold                    &    \CheckmarkBold                   &                            \\
ChatGLM             & Tsinghua         & 6B                    & Weights         &  \CheckmarkBold                    &    \CheckmarkBold                   &                            \\
Alpaca              & Stanford         & 7B                    & Weights         &    \CheckmarkBold                  &   \CheckmarkBold                    &                            \\
MOSS                & Fudan            & 16B                   & Weights         &  \CheckmarkBold                    &    \CheckmarkBold                   &                            \\
LLaMa               & Meta             & 7B, 13B               & Weights         & \CheckmarkBold                     &    \CheckmarkBold                   &                            \\ \bottomrule
\end{tabular}
\caption{Models evaluated in this paper. The ``access" columns show whether we have full access to the model weights or we can only access through API or web. SD stands for Static Data, DD stands for Dynamic Data, and ED stands for Experimental Data. Marking ``\CheckmarkBold" means we evaluate the corresponding model on this subset.}
\label{tab:models}
\end{table*}

\begin{figure}[H]
    \centering
    \includegraphics[width=0.40\textwidth]{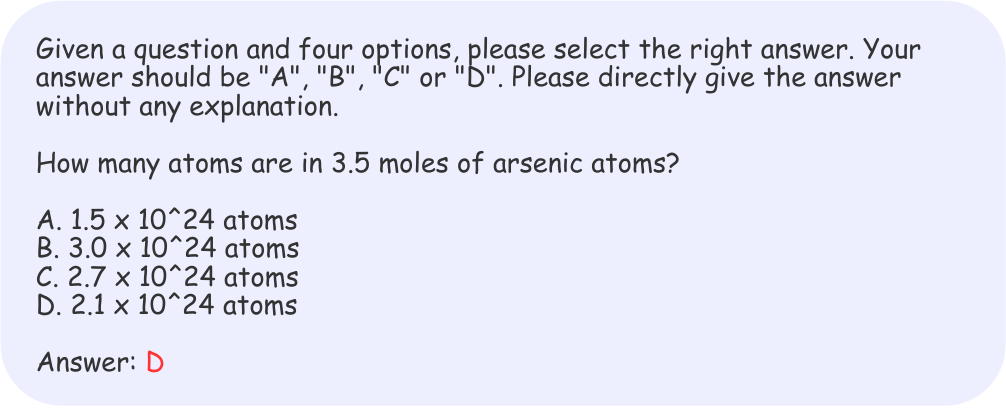}
    \caption{An example of the prompt we used for AO setting. The red text is the response from the model, while the black text is the inputted prompt.}
    \label{fig:prompt}
\end{figure}

\begin{figure}[H]
    \centering
    \includegraphics[width=0.40\textwidth]{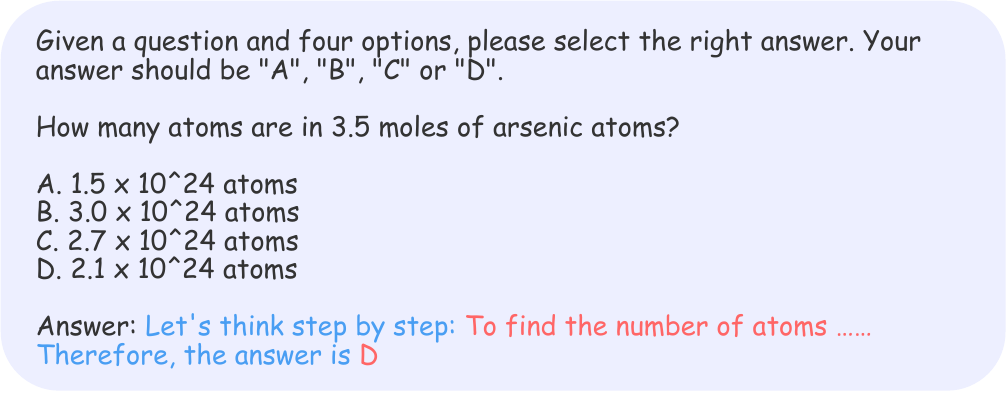}
    \caption{An example of the prompt we used for CoT setting. The red text is the response from the model, while the blue text and black text are the inputted prompt.}
    \label{fig:cot}
\end{figure}

\begin{table*}[!h]
\centering
\begin{tabular}{l|cccc|ccc|c|c}
\toprule
\multirow{2}{*}{\textbf{Model}} & \multicolumn{4}{c|}{\textbf{Static Data}}                         & \multicolumn{3}{c|}{\textbf{Chemistry(DD)}}       & \textbf{Physics(DD)} & \textbf{Exp} \\
                                & Biology        & Chemistry      & Physics        & Avg.           & Acc.           & BLEU           & MSE             & Acc.                 & Score                 \\ \midrule
GPT-4                           & \textbf{84.49} & \textbf{69.38} & \textbf{65.22} & \textbf{73.93} & \textbf{11.05} & \textbf{23.78} & 891.09          & 25.84                & \textbf{93.31}        \\
GPT-3.5-turbo                   & 76.42          & 64.30          & 52.30          & 66.97          & 7.65           & 18.86          & 2008.72         & 21.80                & 88.27                 \\
Claude-v1.3                     & 72.58          & 59.72          & 54.94          & 63.45          & 5.75           & 21.98          & 1489.87         & 26.14                & 85.73                 \\
Claude-instant-v1.1             & 70.43          & 53.36          & 52.30          & 58.92          & 0.45           & 16.07          & 8258.46         & 21.46                & 87.50                 \\
Galactica-30B                   & 66.48          & 50.16          & 44.65          & 54.96          & 0.9            & 4.14           & \textbf{485.99} & 22.47                & -          \\
Vicuna-13B                      & 58.39          & 53.06          & 45.13          & 53.93          & 0.95           & 6.50           & 766.64          & 21.24                & -                     \\
Galactica-6.7B                  & 57.84          & 50.77          & 30.99          & 50.87          & 1.55           & 6.47           & 5519.82         & 20.79                & -                     \\
ChatGLM2-6B                     & 58.62          & 44.00          & 40.26          & 48.44          & 0.2            & 1.86           & 3449.44         & 24.83                & -                     \\
ChatGLM-6B                      & 52.54          & 45.36          & 40.80          & 47.23          & 0.75           & 2.44           & 10303.90        & 21.01                & -                     \\
Alpaca-7B                       & 56.66          & 42.43          & 37.01          & 46.54          & 0.2            & 2.92           & 428419.27       & 26.74                & -                     \\
MOSS-16B                        & 47.71          & 33.87          & 31.73          & 38.23          & 0.1            & 7.37           & 30505.17        & 24.27                & -                     \\
LLaMa-13B                       & 48.59          & 33.56          & 19.48          & 36.96          & 0.3            & 5.21           & 3707.01         & 7.08                 & -                     \\
LLaMa-7B                        & 36.24          & 26.38          & 15.02          & 28.37          & 0.5            & 1.26           & 11305.65        & 14.38                & -                     \\
ERNIE Bot                       & -              & -              & -              & -              & -              & -              & -               & -                    & 61.12                 \\
SparkDesk                       & -              & -              & -              & -              & -              & -              & -               & -                    & 33.69                 \\ \bottomrule
\end{tabular}
\caption{Model performances of Answer-Only setting. The leaderboard is sorted by the average accuracy of Static Data.}
\label{tab:ao_res}
\end{table*}

\subsection{Experiment Setup}

\paragraph{Prompts} We evaluate LLMs in both Answer-Only (AO) and Chain-Of-Thought (CoT)~\cite{kojima2022large} settings. The prompts we used are shown in Figures \ref{fig:prompt} and \ref{fig:cot} respectively. Furthermore, we also evaluate using 3-shot setting, where the three exemplars are selected from the dev set.

\begin{figure*}[h]
    \centering
    \includegraphics[width=0.86\textwidth]{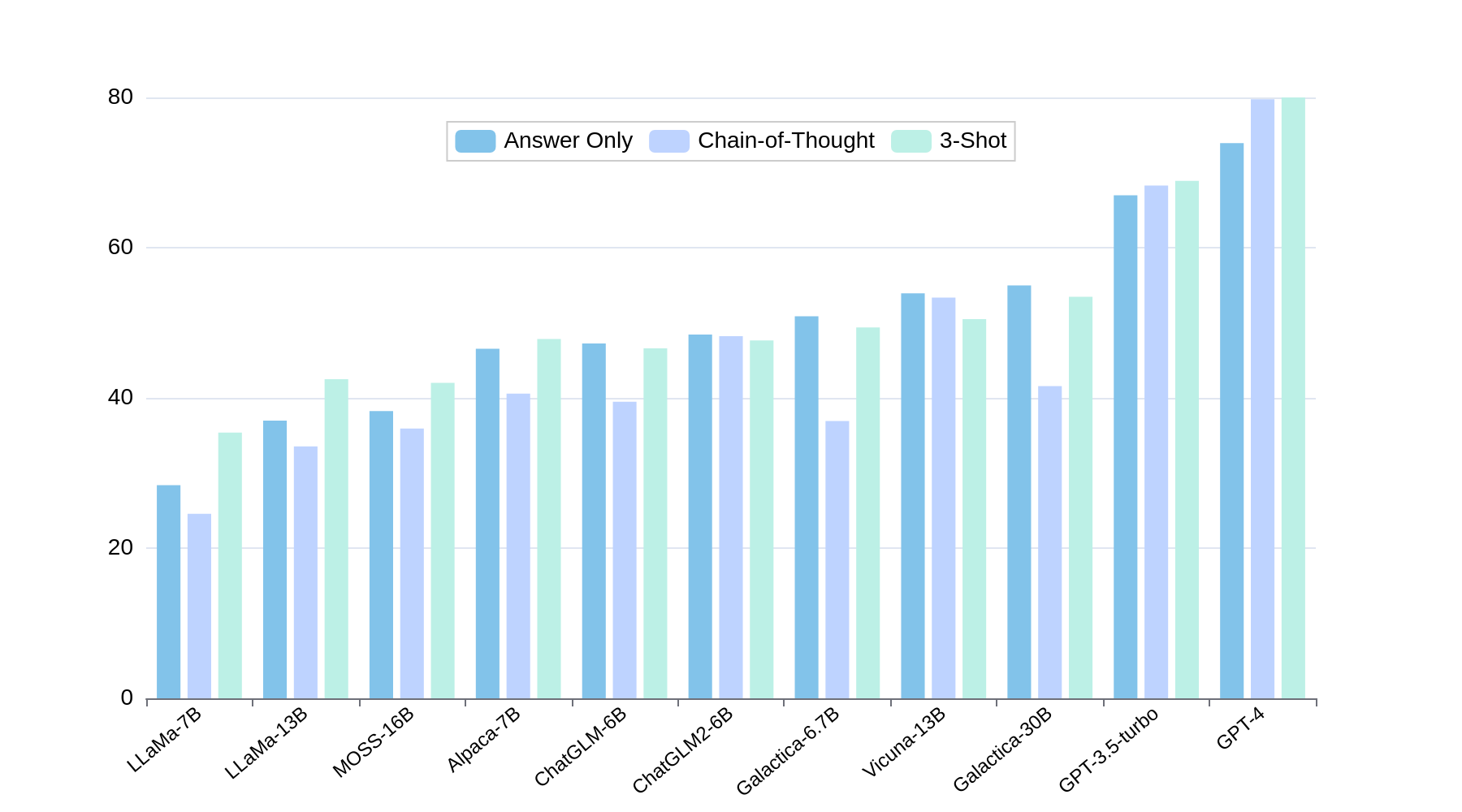}
    \caption{Accuracy on Answer Only, Chain-of-Thought and 3-Shot settings of each LLMs for Static Data.}
    \label{fig:res_compare}
\end{figure*}

\paragraph{Models} In order to comprehensively assess the scientific capabilities of Large Language Models (LLMs), we evaluate 15 high-performing LLMs that are widely accessible. These models are selected to represent a diverse range of organizations and vary in size. The details of these models are summarized in Table \ref{tab:models}.

\begin{itemize}
    \item GPT-3.5-turbo and GPT-4~\cite{schulman2022chatgpt,openai2023gpt4} are the strongest GPT model variants from OpenAI that have undergone pretraining, instruction tuning, and reinforcement learning from human feedback (RLHF,~\cite{ouyang2022training}).
    \item Claude\footnote{\url{https://www.anthropic.com/index/introducing-claude.}}, developed by Anthropic, is often considered comparable to GPT-3.5-turbo. We evaluate both the Claude-v1.3 and Claude-instant-v1.1, a lighter version.
    \item ERNIE Bot\footnote{\url{https://yiyan.baidu.com/}} is developed by Baidu, possessing deep semantic understanding and generation capabilities across modalities and languages. SparkDesk\footnote{\url{https://xinghuo.xfyun.cn/}} is proposed by iFLYTEK. It has cross-domain knowledge and language understanding capabilities and can understand and execute tasks based on natural dialogue.
    \item LLaMa~\cite{touvron2023llama}, developed by Meta, is probably the best open-weight foundation model so far. Vicuna~\cite{zheng2023judging} and Alpaca~\cite{taori2023stanford} are both fine-turned from LLaMa with supervised instruction fine-tuning. It is believed that the performance of Vicuna is better than that of Alpaca.
    \item Galactica~\cite{GALACTICA} is also developed by Meta, which is trained on a large-scale scientific corpus. It is developed to study the use of language models for the automatic organization of science and can perform numerous scientific tasks, such as citation prediction, scientific QA, and molecular property prediction.
    \item ChatGLM and ChatGLM2, created by Tsinghua University, are based on GLM architecture~\cite{du2022glm}, and further adapted on conversational data. MOSS~\cite{sun2023moss}, developed by Fudan University, is the first publicly available Chinese LLM, and it follows a training procedure similar to ChatGPT.
\end{itemize}

We evaluate GPT-3.5-turbo, GPT4 and Claude on all three subsets, including Static Data, Dynamic Data, and Experimental Data. Since we can only assess ERNIE Bot and SparkDesk through web interface, we evaluate these two models only on the Experimental Data. And for the rest LLMs with billions or tens of billions of parameters, since the length of the Experimental Data exceeds the length limit of these models\footnote{The maximum context length of ChatGLM2 is extended to 32k, while it has limited ability to understand long texts.}, we evaluate them on Static Data and Dynamic Data, as is shown in Table \ref{tab:models}.

\paragraph{Evaluation Metrics} 

In the case of Static Data, all questions are objective, making accuracy the appropriate evaluation metric. For Dynamic Data, the physics questions are presented as multiple-choice questions, which can also be evaluated using accuracy. Conversely, the chemistry questions involve complex components, such as ``What is the molecular weight of A?" and ``What is the SMILES expression of B?". Hence, for questions with numerical answers, we employ MSE\footnote{If the predictions do not contain any number, we will regard the MSE as $1 \times 10^{10}$} as the evaluation metric, while for questions with string answers, we utilize the BELU score~\cite{papineni2002bleu}. Additionally, we also calculate the extract match scores. As for Experimental Data, each experiment consists of multiple open-ended questions. As a result, we assess the model-generated responses manually.

\subsection{Experiment Results}

\begin{table*}[h]
\centering
\begin{tabular}{l|ccc|ccc}
\toprule
\multirow{2}{*}{\textbf{Model}} & \multicolumn{3}{c|}{\textbf{Chemistry}} & \multicolumn{3}{c}{\textbf{Physics}} \\
                                & AO           & CoT             & 3-Shot              & AO               & CoT      & 3-Shot            \\ \midrule
GPT-4                           & \textbf{11.05}  & \textbf{11.65}  $\uparrow$  & \textbf{12.42}$\uparrow$   & 25.84                       & 17.98 $\downarrow$                         & \textbf{51.01} $\uparrow$    \\
GPT-3.5-turbo                   & 7.65    & 10.20 $\uparrow$     & 8.85 $\uparrow$       & 21.80                                & \textbf{47.19} $\uparrow$                 & 25.39 $\sim$             \\
Galactica-6.7B                  & 1.55      & 1.75 $\uparrow$     & 3.05 $\uparrow$        & 20.79                               & 23.37 $\sim$                         & 21.12 $\sim$             \\
Vicuna-13B                      & 0.95                  & 1.95 $\uparrow$      & 1.80 $\uparrow$                            & 21.24    & 18.65 $\sim$                       & 23.37$\sim$              \\
Galactica-30B                   & 0.90         & 2.60 $\uparrow$       & 3.30  $\uparrow$          & 22.47                         & 14.72 $\downarrow$                      & 22.58  $\sim$            \\
ChatGLM-6B                      & 0.75      & 0.80  $\uparrow$        & 1.15 $\uparrow$         & 21.01                            & 25.39 $\sim$                       & 23.37 $\sim$             \\
LLaMa-7B                        & 0.50        & 0.10 $\downarrow$   & 1.55 $\uparrow$       & 18.65                                & 9.66  $\downarrow$                         & 27.53 $\uparrow$             \\
LLaMa-13B                       & 0.30        & 0.25 $\sim$   & 2.11 $\uparrow$           & 7.08                               & 5.84  $\sim$                       & 22.70 $\uparrow$             \\
ChatGLM2-6B                     & 0.20       & 2.65 $\uparrow$           & 1.60 $\uparrow$        & 24.83                          & 25.39 $\sim$                       & 26.74 $\sim$             \\
Alpaca-7B                       & 0.20         & 0.65 $\uparrow$      & 2.10 $\uparrow$        & \textbf{26.71}                    & 28.43 $\sim$                       & 25.62 $\sim$             \\
MOSS-16B                        & 0.10    & 0.85 $\uparrow$      & 0.65 $\uparrow$               & 24.27                            & 25.06  $\sim$                     & 26.40   $\sim$           \\ \bottomrule
\end{tabular}
\caption{Results on Answer-Only, Chain-of-Thought and 3-Shot settings of each LLM for Dynamic Data. $\uparrow$ means the performance is slightly better than that under Answer-Only setting, $\downarrow$ means worse, and $\sim$ means the performance is nearly the same.}
\label{tab:res_compare_dd}
\end{table*}

\subsubsection{Answer-Only Setting}
Answer-only results of all the models on the test set are shown in Table \ref{tab:ao_res} and detailed results of Static Data across different knowledge domains are provided in Appendix B. Analyzing the results of Static Data, GPT-4 demonstrates significantly superior performance compared to other models. And only GPT-4, GPT-3.5-turbo, and Claude-v1.3 achieve an average accuracy exceeding 60\%, which highlights the challenge posed by SciEval. 

For the results of Dynamic Data, GPT-4 performs the best in terms of average accuracy and BLEU score. However, for counting and calculation questions, Galactica-30B yields the best results, indicating its strong aptitude in the field of science. Conversely, models with billions or tens of billions of parameters perform poorly on the chemistry subset, suggesting their limited knowledge about molecules. Regarding the performance of models on the physics subset, since all questions are 4-choices questions, the accuracy should be at least 25\%. However, none of these models achieve satisfactory results in this subset.

As for Experimental Data, GPT-series models and Claude-series models achieve good results, while the other two models are not. The detailed scores models reached in each experiment are shown in Appendix C. However, although some models could get a great performance, during experiments, we find that these models are good at experimental principles and designing, while when it comes to analyzing the experiment results, the performances are not satisfying.

\subsubsection{CoT Setting and 3-Shot setting} Comparison of experiment results among Answer-Only, Chain-of-Thought and 3-Shot settings are shown in Figure \ref{fig:res_compare} and Table \ref{tab:res_compare_dd}.\footnote{When evaluating on CoT and 3-Shot settings, Claude-Instant and Claude are not available for us, due to the limitation of API.} And we refer detailed results to Appendix A and B.

The experimental results from Static Data reveal that solely the GPT-series LLMs get performance enhancement within the CoT setting due to the limited CoT capabilities of other LLMs. As for the 3-Shot setting, roughly half of the LLMs analyzed demonstrate superior performances relative to the Answer-Only setting. The performances of the remaining LLMs are closely similar to those observed within the Answer-Only setting.

From the experimental results of Dynamic Data, it is observed that both CoT and 3-Shot significantly enhance the performance of most Language Learning Models (LLMs) in the chemistry subset. However, the performances achieved are still not up to the mark. In the physics subset, the impact of CoT and 3-Shot on most LLMs is less pronounced, resulting in nearly random performances. Under the CoT setting, GPT-3.5-turbo achieves an accuracy of 47.19, suggesting a robust understanding of physical principles. Conversely, the performance of GPT-4 is markedly poor, from which we find that despite its extensive knowledge of physical principles, it frequently employs incorrect formulas to solve problems. Nevertheless, GPT-4 attains an accuracy of 51.01 under 3-Shot setting, the highest among all models, demonstrating its ability to learn from a mere three examples.

\subsection{Discussion}

\subsubsection{Training on large-scale scientific corpus is helpful.}
Based on experimental results (Table \ref{tab:ao_res}), Galactica~\cite{GALACTICA}, which has been trained on an extensive scientific corpus, significantly outperforms other LLMs with a comparable number of parameters, although Galactica is trained with a much smaller amount of data. Remarkably, when tested on Dynamic Data, Galactica surpasses the GPT-series and Claude-series LLMs in computational problems.

\subsubsection{Most LLMs perform bad on calculation problems, especially in physics domain.} 
Detailed results across various knowledge domains on Static Data (refer to Appendix B) reveal that most LLMs underperform in the Scientific Calculation domain, while demonstrate relatively superior performance in other domains, which is particularly acute in the field of physics. Similar issues are also observed in Dynamic Data and Experimental Data. In the context of Dynamic Data, the mean square error, employed to evaluate calculation abilities within the chemistry subset, is exceedingly high for most LLMs, and almost all LLMs can only achieve nearly random performance within the physics subset. Regarding Experimental Data, our findings indicate that these LLMs struggle with the analysis of experimental results.

\section{Conclusion}

In this paper, we introduce SciEval, a benchmark designed to evaluate scientific capabilities of LLMs. SciEval comprises about 18,000 challenging scientific questions, covering three fundamental fields of science. SciEval assesses the scientific ability of LLMs across four dimensions. It incorporates both objective and subjective questions, and employs dynamic data generation to mitigate potential data leakage. We conduct comprehensive experiments on various advanced LLMs using SciEval and perform thorough analyses. Our experimental results reveal that most LLMs do not perform well on our benchmark, with the exception of the GPT-series and Claude-series LLMs. We hope that SciEval can serve as a robust benchmark for assessing scientific capabilities of LLMs.

\section*{Acknowledgements}

This work is funded by the China NSFC Projects (92370206, U23B2057, 62106142 and 62120106006) and Shanghai Municipal Science and Technology Major Project (2021SHZDZX0102).

\bibliography{aaai24}

\end{document}